\documentclass{article}

\usepackage[final]{nips_2017}

\usepackage[utf8]{inputenc} 
\usepackage[T1]{fontenc}    
\usepackage{url}            
\usepackage{booktabs}       
\usepackage{amsfonts}       
\usepackage{nicefrac}       
\usepackage{microtype}      
\usepackage{color}
\usepackage{slashbox}
\usepackage{graphicx}
\usepackage{algorithm} 
\usepackage{enumerate}
\usepackage{algpseudocode}
\usepackage{array}
\usepackage{amsmath,amssymb,amsthm}
\usepackage{wrapfig}

\DeclareMathOperator*{\argmin}{arg\!\min}

\title{Shallow Updates for Deep Reinforcement Learning}

\author{
Nir Levine$^*$ \\
Dept. of Electrical Engineering\\
The Technion - Israel Institute of Technology\\
Israel, Haifa 3200003 \\
\texttt{levin.nir1@gmail.com} \\
\And
Tom Zahavy$^*$ \\
Dept. of Electrical Engineering\\
The Technion - Israel Institute of Technology\\
Israel, Haifa 3200003 \\
\texttt{tomzahavy@campus.technion.ac.il} \\
\AND
Daniel J. Mankowitz\\
Dept. of Electrical Engineering\\
The Technion - Israel Institute of Technology\\
Israel, Haifa 3200003 \\
\texttt{danielm@tx.technion.ac.il} \\
\And
Aviv Tamar \\
Dept. of Electrical Engineering and\\ Computer Sciences, 
UC Berkeley\\
Berkeley, CA 94720 \\
\texttt{avivt@berkeley.edu} \\
\And
Shie Mannor\\
Dept. of Electrical Engineering\\
The Technion - Israel Institute of Technology\\
Israel, Haifa 3200003 \\
\texttt{shie@ee.technion.ac.il} \\
\AND
* Joint first authors. Ordered alphabetically by first name.
}

\begin{document}

\maketitle

\begin{abstract}
Deep reinforcement learning (DRL) methods such as the Deep Q-Network (DQN) have achieved state-of-the-art results in a variety of challenging, high-dimensional domains. This success is mainly attributed to the power of deep neural networks to learn rich domain representations for approximating the value function or policy.  Batch reinforcement learning methods with linear representations, on the other hand, are more stable and require less hyper parameter tuning. Yet, substantial feature engineering is necessary to achieve good results. In this work we propose a hybrid approach -- the Least Squares Deep Q-Network (LS-DQN), which combines rich feature representations learned by a DRL algorithm with the stability of a linear least squares method. We do this by periodically re-training the last hidden layer of a DRL network with a batch least squares update.
Key to our approach is a Bayesian regularization term for the least squares update, which prevents over-fitting to the more recent data. We tested LS-DQN on five Atari games and demonstrate significant improvement over vanilla DQN and Double-DQN. We also investigated the reasons for the superior performance of our method. Interestingly, we found that the performance improvement can be attributed to the large batch size used by the LS method when optimizing the last layer.

\end{abstract}

\section{Introduction}
\label{sec:intro}
Reinforcement learning (RL) is a field of research that uses dynamic programing (DP; \citealt{Bertsekas2008}), among other approaches, to solve sequential decision making problems. The main challenge in applying DP to real world problems is an exponential growth of computational requirements as the problem size increases, known as the curse of dimensionality \citep{Bertsekas2008}. 
 
RL tackles the curse of dimensionality by \emph{approximating} terms in the DP calculation such as the value function or policy. Popular function approximators for this task include deep neural networks, henceforth termed deep RL (DRL), and linear architectures, henceforth termed shallow RL (SRL). 

SRL methods have enjoyed wide popularity over the years (see, e.g.,~ \citealt{tsitsiklis1997analysis,Bertsekas2008} for extensive reviews). 
%
%
In particular, batch algorithms based on a least squares (LS) approach, such as Least Squares Temporal Difference (LSTD, \citealt{Lagoudakis2003}) and Fitted-Q Iteration (FQI, \citealt{ernst2005tree}) are known to be stable and data efficient. However, the success of these algorithms crucially depends on the quality of the feature representation. Ideally, the representation encodes rich, expressive features that can accurately represent the value function. However, in practice, finding such good features is difficult and often hampers the usage of linear function approximation methods. 

In DRL, on the other hand, the features are learned \emph{together with the value function} in a deep architecture. Recent advancements in DRL using convolutional neural networks demonstrated learning of expressive features \citep{zahavy2016graying,wang2015dueling} and state-of-the-art performance in challenging tasks such as video games (\citealt{Mnih2015,tessler2016deep,Mnih2016asynchronous}), and Go \citep{silver2016mastering}. To date, the most impressive DRL results (E.g., the works of \citealt{Mnih2015}, \citealt{Mnih2016asynchronous}) were obtained using \emph{online} RL algorithms, based on a stochastic gradient descent (SGD) procedure.
%
%
%
%

On the one hand, SRL is stable and data efficient. On the other hand, DRL learns powerful representations. This motivates us to ask: can we combine DRL with SRL to leverage the benefits of both?
%
%

In this work, we develop a \emph{hybrid approach} that combines batch SRL algorithms with online DRL. Our main insight is that the last layer in a deep architecture can be seen as a linear representation, with the preceding layers encoding features. Therefore, the last layer can be learned using standard SRL algorithms.
Following this insight, we propose a method that repeatedly \emph{re-trains the last hidden layer} of a DRL network with a batch SRL algorithm, using data collected throughout the DRL run. 

We focus on value-based DRL algorithms (e.g., the popular DQN of \citealt{Mnih2015}) and on SRL based on LS methods\footnote{Our approach can be generalized to other DRL/SRL variants.}
, and propose the Least Squares DQN algorithm (LS-DQN). Key to our approach is a novel regularization term for the least squares method that uses the DRL solution as a prior in a Bayesian least squares formulation. Our experiments demonstrate that this hybrid approach significantly improves performance on the Atari benchmark for several combinations of DRL and SRL methods. 

To support our results, we performed an in-depth analysis to tease out the factors that make our hybrid approach outperform DRL. Interestingly, we found that the improved performance is mainly due to the large batch size of SRL methods compared to the small batch size that is typical for DRL.

\section{Background}\label{sec:background}
In this section we describe our RL framework and several shallow and deep RL algorithms that will be used throughout the paper.

\textbf{RL Framework:} We consider a standard RL formulation \citep{Sutton1998} based on a Markov Decision Process (MDP). An MDP is a tuple $\langle S,A,R,P,\gamma \rangle$, where $S$ is a finite set of states, $A$ is a finite set of actions, and $\gamma \in [0,1]$ is the discount factor. A transition probability function $P:S\times A \rightarrow \Delta_{S}$ maps states and actions to a probability distribution over next states. Finally, $R:S \times A \rightarrow [R_{min}, R_{max}]$ denotes the reward. The goal in RL is to learn a policy $\pi:S \rightarrow \Delta_A$ that solves the MDP by maximizing the expected discounted return $\mathbb{E} \left[ \left. \sum_{t=0}^\infty \gamma^t r_t \right| \pi \right]$. Value based RL methods make use of the action value function $Q^{\pi}(s,a)=\mathbb{E}[\sum_{t=0}^\infty \gamma^t r_t \vert s_t = s, a_t = a, \pi ]$, which represents the expected discounted return of executing action $a \in A$ from state $s \in S$ and following the policy $\pi$ thereafter. The optimal action value function $Q^*(s,a)$ obeys a fundamental recursion known as the Bellman equation $Q^*(s,a) = \mathbb{E}\left[ \left. r_t + \gamma \max_{a'} Q^*(s_{t+1}, a') \right| s_t=s, a_t=a \right]$.
\subsection{SRL algorithms}
\textbf{Least Squares Temporal Difference Q-Learning (LSTD-Q):} LSTD \citep{Barto1996} and LSTD-Q \citep{Lagoudakis2003} are batch SRL  algorithms. LSTD-Q 
learns a control policy $\pi$ from a batch of samples
by estimating a linear approximation $\hat{Q}^{\pi} = \Phi w^\pi$ of the action value function $Q^{\pi}\in \mathbb{R}^{|S||A|}$, 
where $w^\pi \in \mathbb{R}^k$ are a set of weights and $\Phi \in \mathbb{R}^{|S||A|\times k}$ is a feature matrix. Each row of $\Phi$ represents a feature vector for a state-action pair $\langle s,a \rangle$. The weights $w^\pi$ are learned by enforcing 
$\hat{Q}^{\pi}$ to satisfy a fixed point equation w.r.t.~
the projected Bellman operator,
%
%
resulting in a system of linear equations 
$A w^\pi = b$, where $A = \Phi^T (\Phi - \gamma \mathbf{P} \Pi_\pi \Phi)$ and $b=\Phi^T \mathcal{R}$. Here, $\mathcal{R} \in \mathbb{R}^{|S||A|}$ is the reward vector, $\mathbf{P} \in \mathbb{R}^{|S||A| \times |S|}$ is the transition matrix and $\Pi_\pi \in \mathbb{R}^{|S|\times |S||A|}$ is a matrix describing the policy. Given a set of $N_{SRL}$ samples $D = \{s_i,a_i,r_i,s_{i+1}\}_{i=1}^{N_{SRL}}$, we can approximate $A$ and $b$ with the following empirical averages:
\begin{equation}
\label{eq:lstdq}
\tilde{A} = \frac{1}{N_{SRL}} \sum_{i=1}^{N_{SRL}} \biggr[ \phi(s_i,a_i)^T \biggl(\phi(s_i,a_i) - \gamma \phi(s_{i+1},\pi(s_{i+1}))\biggr) \biggr], \enspace
\tilde{b} = \frac{1}{N_{SRL}} \sum_{i=1}^{N_{SRL}} \biggl[\phi(s_i,a_i)^T r_i \biggr]. 
\end{equation}
The weights $w^{\pi}$ can be calculated using a least squares minimization:
$\textstyle \tilde{w}^\pi = \arg \min_w \Vert \tilde{A}w - \tilde{b} \Vert_{2}^2$
or by calculating the pseudo-inverse:
$\tilde{w}^\pi = \tilde{A}^{\dagger} \tilde{b}$.
%
%
LSTD-Q is an \emph{off-policy} algorithm:
the same set of samples $D$ can be used to train any policy $\pi$ so long as $\pi(s_{i+1})$ is defined for every $s_{i+1}$ in the set.

\textbf{Fitted Q Iteration (FQI):}
The FQI algorithm \citep{ernst2005tree} is a batch SRL algorithm that computes iterative approximations of the Q-function using regression. 
At iteration $N$ of the algorithm, the set $D$ defined above and the approximation from the previous iteration $Q^{N-1}$ are used to generate 
supervised learning targets:  ${y_i = r_i+\gamma \max_{a'} Q^{N-1}(s_{i+1},a^{'}),\quad,\forall i\in N_{SRL}}$. These targets are then used by a supervised learning (regression) method to compute the next function in the sequence $Q^N$, by minimizing the MSE loss ${Q^N = \argmin_Q \sum_{i=1}^{N_{SRL}} (Q(s_i, a_i) - (r_i + \gamma \max_{a'} Q^{N-1}(s_{i+1},a')))^2}$. For a linear function approximation $Q_n(a,s) = \phi ^T (s,a) w_n$, LS can be used to give the FQI solution
${\textstyle w_n = \arg \min_w \Vert \tilde{A}w - \tilde{b} \Vert_{2}^2, }$
 where $\tilde{A},\tilde{b}$ are given by:
\begin{equation}
\label{eq:fqi}
\tilde{A} = \frac{1}{N_{SRL}} \sum_{i=1}^{N_{SRL}} \biggr[ \phi(s_i,a_i)^T \phi(s_i,a_i) \biggr], \qquad \tilde{b} = \frac{1}{N_{SRL}} \sum_{i=1}^{N_{SRL}} \biggl[\phi(s_i,a_i) ^T y_i \biggr] \enspace .
\end{equation}

The FQI algorithm can also be used with non-linear function approximations such as trees \citep{ernst2005tree} and neural networks \citep{riedmiller2005neural}. The DQN algorithm \citep{Mnih2015} can be viewed as online form of FQI.

\subsection{DRL algorithms}
\textbf{Deep Q-Network (DQN):}  
The DQN algorithm~\citep{Mnih2015} learns the Q function by minimizing the mean squared error of the Bellman equation, defined as $\mathbb{E}_{s_t,a_t,r_t,s_{t+1}} \Vert Q_\theta(s_t,a_t) - y_t \Vert_{2}^2$, where 
%
%
${y_{t}=r_{t}+\gamma\max_{a'}Q_{\theta_{target}}(s_{t+1},a^{'})}$. The DQN maintains two separate networks, namely the current network with weights $\theta$ and the target network with weights $\theta_{target}$. Fixing the target network makes the DQN algorithm equivalent to \textbf{FQI} (see the FQI MSE loss defined above), where the regression algorithm is chosen to be SGD (RMSPROP, \citealt{hinton2012neural}). The DQN is an off-policy learning algorithm. Therefore, the tuples $\langle s_t, a_t, r_t, s_{t+1} \rangle$ that are used to optimize the network weights are first collected from the agent's experience and are stored in an Experience Replay (ER) buffer \citep{lin1993reinforcement} providing improved stability and performance.

\textbf{Double DQN (DDQN):} DDQN~\citep{van2015deep} is a modification of the DQN algorithm that addresses overly optimistic estimates of the value function. This is achieved by performing action selection with the current network $\theta$ and evaluating the action with the target network, $\theta_{target}$, yielding the DDQN target update $y_t=r_t$ if $s_{t+1}$ is terminal, otherwise $y_t = r_t + \gamma Q_{\theta_{target}}(s_{t+1}, \max_{a} Q_{\theta}(s_{t+1}, a))$.

\section{The LS-DQN Algorithm}
\label{sec:alg}

We now present a hybrid approach for DRL with SRL updates\footnote{Code is available online at \url{https://github.com/Shallow-Updates-for-Deep-RL}}. Our algorithm, the LS-DQN Algorithm, periodically switches between training a DRL network and re-training its last hidden layer using an SRL method. \footnote{We refer the reader to Appendix~B for a diagram of the algorithm.}

We assume that the DRL algorithm uses a deep network for representing the Q function\footnote{The features in the last DQN layer are not action dependent. We generate action-dependent features $\Phi\left(s,a \right)$ by zero-padding to a one-hot state-action feature vector. See Appendix~E for more details.}, where the last layer is linear and fully connected. Such networks have been used extensively in deep RL recently (e.g.,~\citealt{Mnih2015,van2015deep,Mnih2016asynchronous}). In such a representation, the last layer, which approximates the Q function, can be seen as a linear combination of features (the output of the penultimate layer), and we propose to learn more accurate weights for it using SRL.

Explicitly, the LS-DQN algorithm begins by training the weights of a DRL network, $w_k$, using a value-based DRL algorithm for $N_{DRL}$ steps (Line 2). LS-DQN then updates the last hidden layer weights, $w^{last}_k$, by executing LS-UPDATE: retraining the weights using a SRL algorithm with $N_{SRL}$ samples (Line 3). 

The LS-UPDATE consists of the following steps. First, data trajectories $D$ for the batch update are gathered using the current network weights, $w_k$ (Line 7). In practice, the current experience replay can be used and \textbf{no additional samples need to be collected}. The algorithm next generates new features $\Phi\left(s,a \right)$ from the data trajectories using the current DRL network with weights $w_k$. This step guarantees that we do not use samples with inconsistent features, as the ER contains features from 'old' networks weights. Computationally, this step requires running a forward pass of the deep network for every sample in $D$, and can be performed quickly using parallelization.

Once the new features are generated, LS-DQN uses an SRL algorithm to re-calculate the weights of the last hidden layer $w^{last}_k$ (Line 9).\\
While the LS-DQN algorithm is conceptually straightforward, we found that naively running it with off-the-shelf SRL algorithms such as FQI or LSTD resulted in instability and a degradation of the DRL performance. The reason is that the `slow' SGD computation in DRL essentially retains information from older training epochs, while the batch SRL method `forgets' all data but the most recent batch. In the following, we propose a novel regularization method for addressing this issue.

\begin{algorithm}[h] 
\caption{LS-DQN Algorithm} 
\begin{algorithmic}[1] 
\Require ~ $w_0$
\For {\texttt{$k=1 \cdots SRL_{iters}$}}
\State $w_k \leftarrow \mbox{trainDRLNetwork}(w_{k-1})$ 
\Comment{Train the DRL network for $N_{DRL}$ steps}
\State $w^{last}_{k} \leftarrow$ \texttt{LS-UPDATE($w_k$)}  \Comment{Update the last layer weights with the SRL solution}
\EndFor
\\
\Function{\texttt{LS-UPDATE}}{$w$}
\State $D \leftarrow \mbox{gatherData}(w)$ 
\State $\Phi(s,a) \leftarrow \mbox{generateFeatures}(D,w)$ 
\State $w^{last} \leftarrow \mbox{SRL-Algorithm}(D,\Phi(s,a))$
\State \textbf{return} $w^{last}$
\EndFunction
\end{algorithmic} 
\label{alg:my_alg}
\end{algorithm}
\subsection*{Regularization}

Our goal is to improve the performance of a value-based DRL agent using a batch SRL algorithm. Batch SRL algorithms, however, do not leverage the knowledge that the agent has gained before the most recent batch\footnote{While conceptually, the data batch can include \emph{all} the data seen so far, due to computational limitations, this is not a practical solution in the domains we consider.}. We observed that this issue prevents the use of off-the-shelf implementations of SRL methods in our hybrid LS-DQN algorithm.

To enjoy the benefits of both worlds, that is, a batch algorithm that can use the accumulated knowledge gained by the DRL network,
we introduce a novel Bayesian regularization method for LSTD-Q and FQI
that uses the last hidden layer weights of the DRL network $w^{last}_k$ as a \emph{Bayesian prior} for the SRL algorithm \footnote{The reader is referred to \cite{ghavamzadeh2015bayesian} for an overview on using Bayesian methods in RL.}. 

\textbf{SRL Bayesian Prior Formulation:} We are interested in learning the weights of the last hidden layer ($w^{last}$), using a least squares SRL algorithm. We pursue a Bayesian approach, where the prior weights distribution at iteration $k$ of LS-DQN is given by ${w_{prior} \sim N(w^{last}_k,\lambda ^{-2})}$, and we recall that $w^{last}_k$ are the 
last hidden layer weights of the DRL network at iteration $SRL_{iter}=k$.
The Bayesian solution for the regression problem in the FQI algorithm 
is given by \citep{box2011bayesian}  
$$w^{last}=(\tilde{A}+\lambda I)^{-1}(\tilde{b}+\lambda w^{last}_k) \enspace ,$$
where $\tilde{A}$ and $\tilde{b}$ are given in Equation \ref{eq:fqi}.
A similar regularization can be added to LSTD-Q based on a regularized  fixed point equation \citep{kolter2009regularization}. Full details are in Appendix~A.

\section{Experiments}
In this section, we present experiments showcasing the improved performance attained by our LS-DQN algorithm compared to state-of-the-art DRL methods. Our experiments are divided into three sections.  In Section \ref{subsec_per_exp}, we start by investigating the behavior of SRL algorithms in high dimensional environments. 
We then show results for the LS-DQN on five Atari domains, in Section \ref{sec:atari}, and compare the resulting performance to regular DQN and DDQN agents. Finally, in Section \ref{sec:ablation}, we present an ablative analysis of the LS-DQN algorithm, which clarifies the reasons behind our algorithm's success.
\subsection{SRL Algorithms with High Dimensional Observations}
\label{subsec_per_exp}
In the first set of experiments, we explore how least squares SRL algorithms perform in domains with high dimensional observations. This is an important step before applying a SRL method within the LS-DQN algorithm.
In particular, we focused on answering the following questions: (1) What regularization method to use? (2) How to generate data for the LS algorithm? (3) How many policy improvement iterations to perform?  

To answer these questions, we performed the following procedure: We trained DQN agents on two games from the Arcade Learning Environment (ALE, \citeauthor{Bellemare2013}); namely, Breakout and Qbert, using the vanilla DQN implementation \citep{Mnih2015}. For each DQN run, we (1) \textbf{periodically} \footnote{Every three million DQN steps, referred to as one epoch (out of a total of 50 million steps).} save the current DQN network weights and ER; (2) Use an SRL algorithm (LSTD-Q or FQI) to re-learn the weights of the last layer, and (3) evaluate the resulting DQN network by temporarily replacing the DQN weights with the SRL solution weights. After the evaluation, \textbf{we replace back the original DQN weights} and continue training.

Each evaluation entails $20$ roll-outs \footnote{Each roll-out starts from a new (random) game and follows a policy until the agent loses all of its lives.} with an $\epsilon$-greedy policy (similar to  \citeauthor{Mnih2015}, $\epsilon=0.05$). This periodic evaluation setup allowed us to effectively experiment with the SRL algorithms and obtain clear comparisons with DQN, without waiting for full DQN runs to complete.

\textbf{(1) Regularization:} Experiments with standard SRL methods without any regularization yielded poor results. We found the main reason to be that the matrices used in the SRL solutions (Equations \ref{eq:lstdq} and \ref{eq:fqi}) are ill-conditioned, resulting in instability. One possible explanation stems from the sparseness of the features. The DQN uses ReLU activations \citep{jarrett2009best}, which causes the network to learn sparse feature representations. For example, once the DQN completed training on Breakout, $96\%$ of features were zero.

\begin{figure}[h]
\begin{center}
    \includegraphics[width=\textwidth]{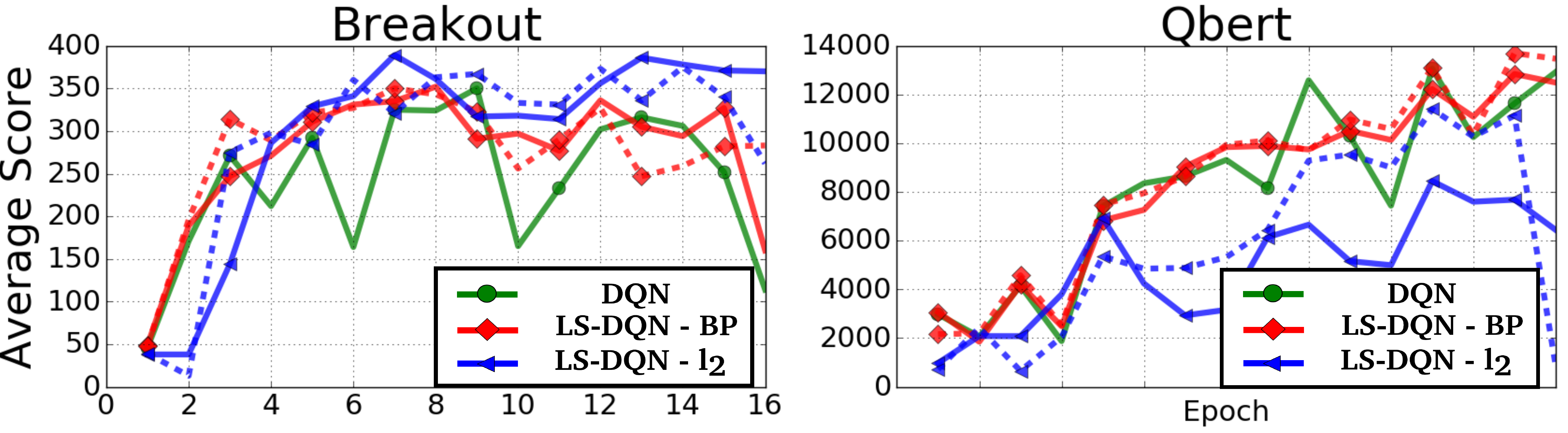}
  \caption{Periodic evaluation for DQN (green), \textbf{LS-DQN$_{\textrm{LSTD-Q}}$} with Bayesian prior regularization (red, solid $\lambda=10$, dashed $\lambda=1$), and $\ell_2$ regularization (blue, solid  $\lambda=0.001$, dashed  $\lambda=0.0001$).}
  \label{fig:reg}
\end{center}
\end{figure}

Once we added a regularization term, we found that the performance of the SRL algorithms improved. We experimented with the $\ell_2$ and Bayesian Prior (BP) regularizers ($\lambda\in\left[0,10^{2}\right]$). While the $\ell_2$ regularizer showed competitive performance in Breakout, we found that the BP performed better across domains (Figure \ref{fig:reg}, best regularizers chosen, shows the average score of each configuration following the explained evaluation procedure, for the different epochs). Moreover, the BP regularizer was not sensitive to the scale of the regularization coefficient. Regularizers in the range $(10^{-1},10^1)$ performed well across all domains. A table of average scores for different coefficients can be found in Appendix~C.1. Note that we do not expect for much improvement as we replace back the original DQN weights after evaluation.\\

\textbf{(2) Data Gathering:} We experimented with two mechanisms for generating data: (1) generating new data from the current policy, and (2) using the ER. We found that the data generation mechanism had a significant impact on the performance of the algorithms. When the data is generated only from the current DQN policy (without ER) the SRL solution resulted in poor performance compared to a solution using the ER (as was observed by \citealt{Mnih2015}). We believe that the main reason the ER works well is that the ER contains data sampled from multiple (past) policies, and therefore exhibits more exploration of the state space.

\textbf{(3) Policy Improvement:} LSTD-Q and FQI are off-policy algorithms and can be applied iteratively on the same dataset (e.g. LSPI, \citealt{Lagoudakis2003}). However, in practice, we found that performing multiple iterations did not improve the results. A possible explanation is that by improving the policy, the policy reaches new areas in the state space that are not represented well in the current ER, and therefore are not approximated well by the SRL solution and the current DRL network.  


\subsection{Atari Experiments}
\label{sec:atari}
We next ran the full LS-DQN algorithm (Alg.~\ref{alg:my_alg}) on five Atari domains: Asterix, Space Invaders, Breakout, Q-Bert and Bowling. We ran the LS-DQN using both DQN and DDQN as the DRL algorithm, and using both LSTD-Q and FQI as the SRL algorithms. We chose to run a LS-update every ${N_{DRL} = 500k}$ steps, for a total of $50$M steps (${SRL_{iters}=100}$). We used the current ER buffer as the `generated' data in the LS-UPDATE function (line~7 in Alg.~\ref{alg:my_alg}, ${N_{SRL}=1M}$), and a regularization coefficient ${\lambda=1}$ for the Bayesian prior solution (both for FQI and LSTQ-Q). We emphasize the we did not use any additional  samples beyond the samples already obtained by the DRL algorithm.

\begin{figure}[h]
\begin{center}
  \includegraphics[width=\textwidth]{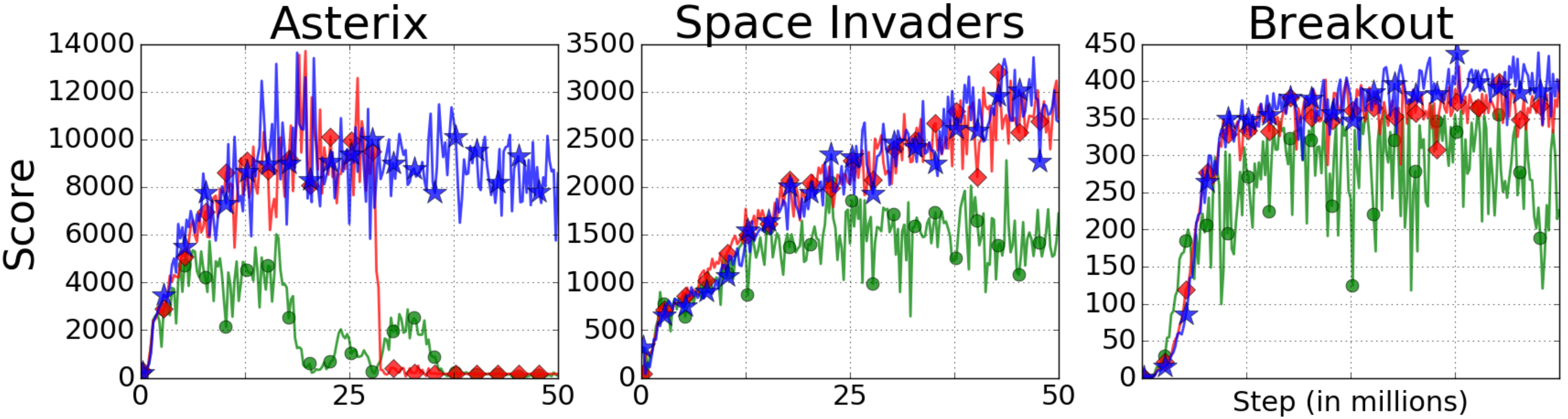}
  \caption{Learning curves of DQN (green), LS-DQN$_{\textrm{LSTD-Q}}$ (red), and LS-DQN$_{\textrm{FQI}}$ (blue).}
  \label{fig:trajs}
\end{center}
\end{figure}

Figure~\ref{fig:trajs} presents the learning curves of the DQN network, LS-DQN with LSTD-Q, and LS-DQN with FQI (referred to as \textbf{DQN}, \textbf{LS-DQN$_{\textrm{LSTD-Q}}$}, and \textbf{LS-DQN$_{\textrm{FQI}}$}, respectively) on three domains: Asterix, Space Invaders and Breakout. Note that we use the same evaluation process as described in \citet{Mnih2015}. We were also interested in a test to measure differences between learning curves, and not only their maximal score. Hence we chose to perform Wilcoxon signed-rank test on the average scores between the three DQN variants. This non-parametric statistical test measures whether related samples differ in their means \citep{wilcoxon1945individual}. We found that the learning curves for both LS-DQN$_{\textrm{LSTD-Q}}$ and LS-DQN$_{\textrm{FQI}}$ were statistically significantly better than those of DQN, with p-values smaller than $1$e-$15$ for all three domains.

Table~\ref{table:max_scores} presents the maximum average scores along the learning curves of the five domains, when the SRL algorithms were incorporated into both DQN agents and DDQN agents (the notation is similar, i.e., \textbf{LS-DDQN$_{\textrm{FQI}}$})\footnote{\label{tab_max} Scores for DQN and DDQN were taken from \cite{van2015deep}.}. 
Our algorithm, LS-DQN, attained better performance compared to the vanilla DQN agents, as seen by the higher scores in Table~\ref{table:max_scores} and Figure~\ref{fig:trajs}. We observe an interesting phenomenon for the game Asterix: In Figure~\ref{fig:trajs}, the DQN's score ``crashes'' to zero (as was observed by \citealt{van2015deep}). LS-DQN$_{\textrm{LSTD-Q}}$ did not manage to resolve this issue, even though it achieved a significantly higher score that that of the DQN. LS-DQN$_{\textrm{FQI}}$, however, maintained steady performance and did not ``crash'' to zero. We found that, in general, incorporating FQI as an SRL algorithm into the DRL agents resulted in improved performance.

\begin{table}[h]
\caption{Maximal average scores across five different Atari domains for each of the DQN variants.}
\begin{center}
\begin{small}
\begin{tabular}{| l || >{\centering\arraybackslash}m{0.1\textwidth} | >{\centering\arraybackslash}m{0.1\textwidth} | >{\centering\arraybackslash}m{0.1\textwidth} | >{\centering\arraybackslash}m{0.1\textwidth} | >{\centering\arraybackslash}m{0.1\textwidth} |} 
\hline
\backslashbox{Algorithm}{Game} & Breakout & Space Invaders & Asterix & Qbert & Bowling \\ \hline \hline
DQN\textsuperscript{\ref{tab_max}} & 401.20 & 1975.50 & 6011.67 & 10595.83 & 42.40 \\ \hline
LS-DQN$_{\textrm{LSTD-Q}}$ & 420.00 & 3207.44 & \textbf{13704.23} & 10767.47 & 61.21\\ \hline
LS-DQN$_{\textrm{FQI}}$ & \textbf{438.55}  & \textbf{3360.81} & 13636.81 & \textbf{12981.42} & \textbf{75.38}\\ \hline \hline
DDQN\textsuperscript{\ref{tab_max}} & 375.00 & 3154.60 & 15150.00 & \textbf{14875.00} & 70.50 \\ \hline
LS-DDQN$_{\textrm{FQI}}$ & \textbf{397.94} &\textbf{4400.83}  & \textbf{16270.45} & 12727.94 & \textbf{80.75}\\ \hline 
\end{tabular}
\label{table:max_scores}
\end{small}
\end{center}
\end{table}

\subsection{Ablative Analysis}
\label{sec:ablation}
In the previous section, we saw that the LS-DQN algorithm has improved performance, compared to the DQN agents, across a number of domains. The goal of this section is to understand the reasons behind the LS-DQN's improved performance by conducting an ablative analysis of our algorithm. For this analysis, we used a DQN agent that was trained on the game of Breakout, in the same manner as described in Section~\ref{subsec_per_exp}. We focus on analyzing the \textbf{LS-DQN$_{\textrm{FQI}}$} algorithm, that has the same optimization objective as DQN (cf. Section \ref{sec:background}), and postulate the following conjectures for its improved performance:
\begin{enumerate}[(i)]
\item The SRL algorithms use a Bayesian regularization term, which is not included in the DQN objective.
\item The SRL algorithms have less hyperparameters to tune and generate an explicit solution compared to SGD-based DRL solutions.
\item Large-batch methods perform better than small-batch methods when combining DRL with SRL.
\item SRL algorithms focus on training the last layer and are easier to optimize.
\end{enumerate}

\textbf{The Experiments:} We started by analyzing the learning method of the last layer (i.e., the `shallow' part of the learning process). We did this by optimizing the last layer, at each LS-UPDATE epoch, using (1) FQI with a Bayesian prior and a LS solution, and (2) an ADAM \citep{kingma2014adam} optimizer with and without an additional Bayesian prior regularization term in the loss function. We compared these approaches for different mini-batch sizes of $32$, $512$, and $4096$ data points, and used $\lambda=1$ for all experiments. 

Relating to conjecture (ii), note that the FQI algorithm has only one hyper-parameter to tune and produces an explicit solution using the whole dataset simultaneously. ADAM, on the other hand, has more hyper-parameters to tune and works on different mini-batch sizes. 

%

\textbf{The Experimental Setup:} The experiments were done in a periodic fashion similar to Section~\ref{subsec_per_exp}, i.e., testing behavior in different epochs over a vanilla DQN run. For both ADAM and FQI, we first collected $80k$ data samples from the ER at each epoch. For ADAM, we performed $20$ iterations over the data, where each iteration consisted of randomly permuting the data, dividing it into mini-batches and optimizing using ADAM over the mini-batches\footnote{\label{C} The selected hyper-parameters used for these experiments can be found in Appendix~D, along with results for one iteration of ADAM.}. We then simulate the agent and report average scores across $20$ trajectories. 

\begin{figure}[h]
\begin{center}
  \includegraphics[width=\textwidth]{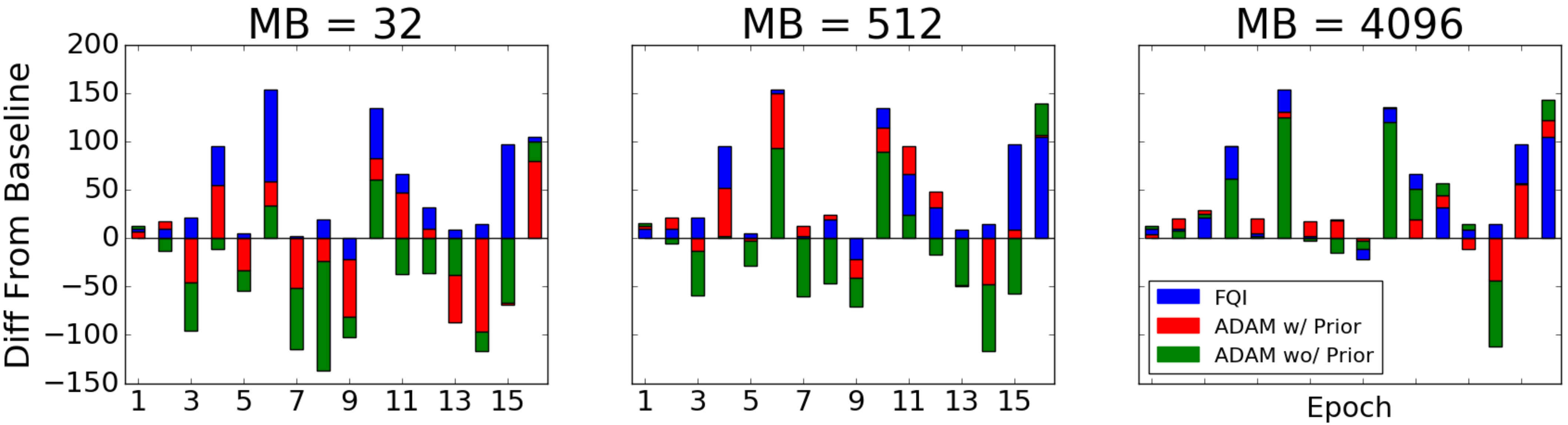}
  \caption{Differences of the average scores, for different learning methods, compared to vanilla DQN. Positive values represent improvement over vanilla DQN. For example, for mini-batch of 32 (left figure), in epoch 3, FQI (blue) out-performed vanilla DQN by 21, while ADAM with prior (red), and ADAM without prior (green) under-performed by -46, and -96, respectively. Note that: (1) as the mini-batch size increases, the improvement of ADAM over DQN becomes closer to the improvement of FQI over DQN, and (2) optimizing the last layer improves performance.}
  \label{fig:score_adam}
\end{center}
\end{figure}

\textbf{The Results:} Figure~\ref{fig:score_adam} depicts the difference between the average scores of (1) and (2) to that of the DQN baseline scores. We see that larger mini-batches result in improved performance. Moreover, the LS solution (FQI) outperforms the ADAM solutions for mini-batch sizes of $32$ and $512$ on most epochs, and even slightly outperforms the best of them (mini-batch size of $4096$ and a Bayesian prior). In addition, a solution with a prior performs better than a solution without a prior.

\textbf{Summary:} Our ablative analysis experiments strongly support conjectures (iii) and (iv) from above, for explaining LS-DQN's improved performance. That is, large-batch methods perform better than small-batch methods when combining DRL with SRL as explained above; and SRL algorithms that focus on training only the last layer are easier to optimize, as we see that optimizing the last layer improved the score across epochs.

We finish this Section with an interesting observation. While the LS solution improves the performance of the DRL agents, we found that the LS solution weights are very close to the baseline DQN solution. See Appendix~D, for the full results. Moreover, the distance was inversely proportional to the performance of the solution. That is, the FQI solution that performed the best, was the closest (in $\ell_{2}$ norm) to the DQN solution, and vice versa. There were orders of magnitude differences between the norms of solutions that performed well and those that did not. Similar results,  i.e., that large-batch solutions find solutions that are close to the baseline, have been reported in \citep{keskar2016large}. We further compare our results with the findings of \citeauthor{keskar2016large} in the section to follow.


\section{Related work}
\label{sec:related}
We now review recent works that are related to this paper.

\textbf{Regularization:} The general idea of applying regularization for feature selection, and to avoid over-fitting is a common theme in machine learning. However, applying it to RL algorithms is challenging due to the fact that these algorithms are based on finding a fixed-point rather than optimizing a loss function  \citep{kolter2009regularization}.
Value-based DRL approaches do not use regularization layers (e.g. pooling, dropout and batch normalization), which are popular in other deep learning methods. The DQN, for example, has a relatively shallow architecture (three convolutional layers, followed by two fully connected layers) without any regularization layers. Recently, regularization was introduced in problems that combine value-based RL with other learning objectives. For example, \cite{hester2017learning} combine RL with supervised learning from expert demonstration, and introduce regularization to avoid over-fitting the expert data; and \cite{kirkpatrick2017overcoming} introduces regularization to avoid catastrophic forgetting in transfer learning. SRL methods, on the other hand, perform well with regularization \citep{kolter2009regularization} and have been shown to converge \cite{farahmand2009regularized}.

\textbf{Batch size:}
Our results suggest that a large batch LS solution for the last layer of a value-based DRL network can significantly improve it's performance. This result is somewhat surprising, as it has been observed by practitioners that using larger batches in deep learning degrades the quality of the model, as measured by its ability to generalize \citep{keskar2016large}. 

However, our method differs from the experiments performed by \citealt{keskar2016large} and therefore does not contradict them, for the following reasons: (1) The LS-DQN Algorithm uses the large batch solution only for the last layer. The lower layers of the network are not affected by the large batch solution and therefore do not converge to a sharp minimum. (2) The experiments of \citep{keskar2016large} were performed for classification tasks, whereas our algorithm is minimizing an MSE loss. (3) \citeauthor{keskar2016large} showed that large-batch solutions work well when piggy-backing (warm-started) on a small-batch solution. Similarly, our algorithm mixes small and large batch solutions as it switches between them periodically. 

Moreover, it was recently observed that flat minima in practical deep learning model classes can be turned into sharp minima via re-parameterization without changing
the generalization gap, and hence it requires further investigation \cite{dinh2017sharp}. In addition, \citeauthor{hoffer2017train} showed that large-batch training can generalize as well as small-batch training by adapting the number of iterations \cite{hoffer2017train}. Thus, we strongly believe that our findings on combining large and small batches in DRL are in agreement with recent results of other deep learning research groups. 

\textbf{Deep and Shallow RL:}  Using the last-hidden layer of a DNN as a feature extractor and learning the last layer with a different algorithm has been addressed before in the literature, e.g., in the context of transfer learning \citep{donahue2014decaf}. In RL, there have been competitive attempts to use SRL with unsupervised features to play Atari \citep{liang2016state,blundell2016model}, but to the best of our knowledge, this is the first attempt that successfully combines DRL with SRL algorithms.

\section{Conclusion}
\label{sec:conclusions}

In this work we presented LS-DQN, a hybrid approach that combines least-squares RL updates within online deep RL. 
LS-DQN obtains the best of both worlds: rich representations from deep RL networks as well as stability and data efficiency of least squares methods. Experiments with two deep RL methods and two least squares methods revealed that a hybrid approach consistently improves over vanilla deep RL in the Atari domain.
Our ablative analysis indicates that the success of the LS-DQN algorithm is due to the large batch updates made possible by using least squares.

This work focused on value-based RL. However, our hybrid linear/deep approach can be extended to other RL methods, such as actor critic \citep{Mnih2016asynchronous}. More broadly, decades of research on linear RL methods have provided methods with strong guarantees, such as approximate linear programming~\citep{desai2012approximate} and modified policy iteration~\citep{scherrer2015approximate}. Our approach shows that with the correct modifications, such as our Bayesian regularization term, linear methods can be combined with deep RL. This opens the door to future combinations of well-understood linear RL with deep representation learning.


\paragraph{Acknowledgement}
This research was supported by the European Community’s Seventh Framework Program (FP7/2007-2013) under grant agreement 306638 (SUPREL). A. Tamar is supported in part by Siemens and the Viterbi Scholarship, Technion.

\newpage
\small
\medskip
\bibliographystyle{icml2016}
\bibliography{rldmbib}

\begin{thebibliography}{34}
\providecommand{\natexlab}[1]{#1}
\providecommand{\url}[1]{\texttt{#1}}
\expandafter\ifx\csname urlstyle\endcsname\relax
  \providecommand{\doi}[1]{doi: #1}\else
  \providecommand{\doi}{doi: \begingroup \urlstyle{rm}\Url}\fi

\bibitem[Barto \& Crites(1996)Barto and Crites]{Barto1996}
Barto, AG and Crites, RH.
\newblock Improving elevator performance using reinforcement learning.
\newblock \emph{Advances in neural information processing systems}, 8:\penalty0
  1017--1023, 1996.

\bibitem[Bellemare et~al.(2013)Bellemare, Naddaf, Veness, and
  Bowling]{Bellemare2013}
Bellemare, Marc~G, Naddaf, Yavar, Veness, Joel, and Bowling, Michael.
\newblock The arcade learning environment: An evaluation platform for general
  agents.
\newblock \emph{Journal of Artificial Intelligence Research}, 47:\penalty0
  253--279, 2013.

\bibitem[Bertsekas(2008)]{Bertsekas2008}
Bertsekas, Dimitri~P.
\newblock Approximate dynamic programming.
\newblock 2008.

\bibitem[Blundell et~al.(2016)Blundell, Uria, Pritzel, Li, Ruderman, Leibo,
  Rae, Wierstra, and Hassabis]{blundell2016model}
Blundell, Charles, Uria, Benigno, Pritzel, Alexander, Li, Yazhe, Ruderman,
  Avraham, Leibo, Joel~Z, Rae, Jack, Wierstra, Daan, and Hassabis, Demis.
\newblock Model-free episodic control.
\newblock \emph{stat}, 1050:\penalty0 14, 2016.

\bibitem[Box \& Tiao(2011)Box and Tiao]{box2011bayesian}
Box, George~EP and Tiao, George~C.
\newblock \emph{Bayesian inference in statistical analysis}.
\newblock John Wiley \& Sons, 2011.

\bibitem[Desai et~al.(2012)Desai, Farias, and Moallemi]{desai2012approximate}
Desai, Vijay~V, Farias, Vivek~F, and Moallemi, Ciamac~C.
\newblock Approximate dynamic programming via a smoothed linear program.
\newblock \emph{Operations Research}, 60\penalty0 (3):\penalty0 655--674, 2012.

\bibitem[Dinh et~al.(2017)Dinh, Pascanu, Bengio, and Bengio]{dinh2017sharp}
Dinh, Laurent, Pascanu, Razvan, Bengio, Samy, and Bengio, Yoshua.
\newblock Sharp minima can generalize for deep nets.
\newblock \emph{arXiv preprint arXiv:1703.04933}, 2017.

\bibitem[Donahue et~al.(2013)Donahue, Jia, Vinyals, Hoffman, Zhang, Tzeng, and
  Darrell]{donahue2014decaf}
Donahue, Jeff, Jia, Yangqing, Vinyals, Oriol, Hoffman, Judy, Zhang, Ning,
  Tzeng, Eric, and Darrell, Trevor.
\newblock Decaf: A deep convolutional activation feature for generic visual
  recognition.
\newblock In \emph{Proceedings of the 30th international conference on machine
  learning (ICML-13)}, pp.\  647--655, 2013.

\bibitem[Ernst et~al.(2005)Ernst, Geurts, and Wehenkel]{ernst2005tree}
Ernst, Damien, Geurts, Pierre, and Wehenkel, Louis.
\newblock Tree-based batch mode reinforcement learning.
\newblock \emph{Journal of Machine Learning Research}, 6\penalty0
  (Apr):\penalty0 503--556, 2005.

\bibitem[Farahmand et~al.(2009)Farahmand, Ghavamzadeh, Mannor, and
  Szepesv{\'a}ri]{farahmand2009regularized}
Farahmand, Amir~M, Ghavamzadeh, Mohammad, Mannor, Shie, and Szepesv{\'a}ri,
  Csaba.
\newblock Regularized policy iteration.
\newblock In \emph{Advances in Neural Information Processing Systems}, pp.\
  441--448, 2009.

\bibitem[Ghavamzadeh et~al.(2015)Ghavamzadeh, Mannor, Pineau, Tamar,
  et~al.]{ghavamzadeh2015bayesian}
Ghavamzadeh, Mohammad, Mannor, Shie, Pineau, Joelle, Tamar, Aviv, et~al.
\newblock Bayesian reinforcement learning: A survey.
\newblock \emph{Foundations and Trends{\textregistered} in Machine Learning},
  8\penalty0 (5-6):\penalty0 359--483, 2015.

\bibitem[Hester et~al.(2017)Hester, Vecerik, Pietquin, Lanctot, Schaul, Piot,
  Sendonaris, Dulac-Arnold, Osband, Agapiou, et~al.]{hester2017learning}
Hester, Todd, Vecerik, Matej, Pietquin, Olivier, Lanctot, Marc, Schaul, Tom,
  Piot, Bilal, Sendonaris, Andrew, Dulac-Arnold, Gabriel, Osband, Ian, Agapiou,
  John, et~al.
\newblock Learning from demonstrations for real world reinforcement learning.
\newblock \emph{arXiv preprint arXiv:1704.03732}, 2017.

\bibitem[Hinton et~al.(2012)Hinton, Srivastava, and Swersky]{hinton2012neural}
Hinton, Geoffrey, Srivastava, NiRsh, and Swersky, Kevin.
\newblock Neural networks for machine learning lecture 6a overview of
  mini--batch gradient descent.
\newblock 2012.

\bibitem[Hoffer et~al.(2017)Hoffer, Hubara, and Soudry]{hoffer2017train}
Hoffer, Elad, Hubara, Itay, and Soudry, Daniel.
\newblock Train longer, generalize better: closing the generalization gap in
  large batch training of neural networks.
\newblock \emph{arXiv preprint arXiv:1705.08741}, 2017.

\bibitem[Jarrett et~al.(2009)Jarrett, Kavukcuoglu, LeCun,
  et~al.]{jarrett2009best}
Jarrett, Kevin, Kavukcuoglu, Koray, LeCun, Yann, et~al.
\newblock What is the best multi-stage architecture for object recognition?
\newblock In \emph{Computer Vision, 2009 IEEE 12th International Conference
  on}, pp.\  2146--2153. IEEE, 2009.

\bibitem[Keskar et~al.(2016)Keskar, Mudigere, Nocedal, Smelyanskiy, and
  Tang]{keskar2016large}
Keskar, Nitish~Shirish, Mudigere, Dheevatsa, Nocedal, Jorge, Smelyanskiy,
  Mikhail, and Tang, Ping Tak~Peter.
\newblock On large-batch training for deep learning: Generalization gap and
  sharp minima.
\newblock \emph{arXiv preprint arXiv:1609.04836}, 2016.

\bibitem[Kingma \& Ba(2014)Kingma and Ba]{kingma2014adam}
Kingma, Diederik and Ba, Jimmy.
\newblock Adam: A method for stochastic optimization.
\newblock \emph{arXiv preprint arXiv:1412.6980}, 2014.

\bibitem[Kirkpatrick et~al.(2017)Kirkpatrick, Pascanu, Rabinowitz, Veness,
  Desjardins, Rusu, Milan, Quan, Ramalho, Grabska-Barwinska,
  et~al.]{kirkpatrick2017overcoming}
Kirkpatrick, James, Pascanu, Razvan, Rabinowitz, Neil, Veness, Joel,
  Desjardins, Guillaume, Rusu, Andrei~A, Milan, Kieran, Quan, John, Ramalho,
  Tiago, Grabska-Barwinska, Agnieszka, et~al.
\newblock Overcoming catastrophic forgetting in neural networks.
\newblock \emph{Proceedings of the National Academy of Sciences}, pp.\
  201611835, 2017.

\bibitem[Kolter \& Ng(2009)Kolter and Ng]{kolter2009regularization}
Kolter, J~Zico and Ng, Andrew~Y.
\newblock Regularization and feature selection in least-squares temporal
  difference learning.
\newblock In \emph{Proceedings of the 26th annual international conference on
  machine learning}. ACM, 2009.

\bibitem[Lagoudakis \& Parr(2003)Lagoudakis and Parr]{Lagoudakis2003}
Lagoudakis, Michail~G and Parr, Ronald.
\newblock Least-squares policy iteration.
\newblock \emph{Journal of machine learning research}, 4\penalty0
  (Dec):\penalty0 1107--1149, 2003.

\bibitem[Liang et~al.(2016)Liang, Machado, Talvitie, and
  Bowling]{liang2016state}
Liang, Yitao, Machado, Marlos~C, Talvitie, Erik, and Bowling, Michael.
\newblock State of the art control of atari games using shallow reinforcement
  learning.
\newblock In \emph{Proceedings of the 2016 International Conference on
  Autonomous Agents \& Multiagent Systems}, 2016.

\bibitem[Lin(1993)]{lin1993reinforcement}
Lin, Long-Ji.
\newblock Reinforcement learning for robots using neural networks.
\newblock 1993.

\bibitem[Mnih et~al.(2015)Mnih, Kavukcuoglu, Silver, Rusu, Veness, Bellemare,
  Graves, Riedmiller, Fidjeland, Ostrovski, et~al.]{Mnih2015}
Mnih, Volodymyr, Kavukcuoglu, Koray, Silver, David, Rusu, Andrei~A, Veness,
  Joel, Bellemare, Marc~G, Graves, Alex, Riedmiller, Martin, Fidjeland,
  Andreas~K, Ostrovski, Georg, et~al.
\newblock Human-level control through deep reinforcement learning.
\newblock \emph{Nature}, 518\penalty0 (7540):\penalty0 529--533, 2015.

\bibitem[Mnih et~al.(2016)Mnih, Badia, Mirza, Graves, Lillicrap, Harley,
  Silver, and Kavukcuoglu]{Mnih2016asynchronous}
Mnih, Volodymyr, Badia, Adria~Puigdomenech, Mirza, Mehdi, Graves, Alex,
  Lillicrap, Timothy~P, Harley, Tim, Silver, David, and Kavukcuoglu, Koray.
\newblock Asynchronous methods for deep reinforcement learning.
\newblock In \emph{International Conference on Machine Learning}, pp.\
  1928--1937, 2016.

\bibitem[Riedmiller(2005)]{riedmiller2005neural}
Riedmiller, Martin.
\newblock Neural fitted q iteration--first experiences with a data efficient
  neural reinforcement learning method.
\newblock In \emph{European Conference on Machine Learning}, pp.\  317--328.
  Springer, 2005.

\bibitem[Scherrer et~al.(2015)Scherrer, Ghavamzadeh, Gabillon, Lesner, and
  Geist]{scherrer2015approximate}
Scherrer, Bruno, Ghavamzadeh, Mohammad, Gabillon, Victor, Lesner, Boris, and
  Geist, Matthieu.
\newblock Approximate modified policy iteration and its application to the game
  of tetris.
\newblock \emph{Journal of Machine Learning Research}, 16:\penalty0 1629--1676,
  2015.

\bibitem[Silver et~al.(2016)Silver, Huang, Maddison, Guez, Sifre, Van
  Den~Driessche, Schrittwieser, Antonoglou, Panneershelvam, Lanctot,
  et~al.]{silver2016mastering}
Silver, David, Huang, Aja, Maddison, Chris~J, Guez, Arthur, Sifre, Laurent, Van
  Den~Driessche, George, Schrittwieser, Julian, Antonoglou, Ioannis,
  Panneershelvam, Veda, Lanctot, Marc, et~al.
\newblock Mastering the game of go with deep neural networks and tree search.
\newblock \emph{Nature}, 529\penalty0 (7587):\penalty0 484--489, 2016.

\bibitem[Sutton \& Barto(1998)Sutton and Barto]{Sutton1998}
Sutton, Richard and Barto, Andrew.
\newblock \emph{{Reinforcement Learning: An Introduction}}.
\newblock MIT Press, 1998.

\bibitem[Tessler et~al.(2017)Tessler, Givony, Zahavy, Mankowitz, and
  Mannor]{tessler2016deep}
Tessler, Chen, Givony, Shahar, Zahavy, Tom, Mankowitz, Daniel~J, and Mannor,
  Shie.
\newblock A deep hierarchical approach to lifelong learning in minecraft.
\newblock \emph{Proceedings of the National Conference on Artificial
  Intelligence (AAAI)}, 2017.

\bibitem[Tsitsiklis et~al.(1997)Tsitsiklis, Van~Roy,
  et~al.]{tsitsiklis1997analysis}
Tsitsiklis, John~N, Van~Roy, Benjamin, et~al.
\newblock An analysis of temporal-difference learning with function
  approximation.
\newblock \emph{IEEE transactions on automatic control 42.5}, pp.\  674--690,
  1997.

\bibitem[Van~Hasselt et~al.(2016)Van~Hasselt, Guez, and Silver]{van2015deep}
Van~Hasselt, Hado, Guez, Arthur, and Silver, David.
\newblock Deep reinforcement learning with double q-learning.
\newblock \emph{Proceedings of the National Conference on Artificial
  Intelligence (AAAI)}, 2016.

\bibitem[Wang et~al.(2016)Wang, Schaul, Hessel, van Hasselt, Lanctot, and
  de~Freitas]{wang2015dueling}
Wang, Ziyu, Schaul, Tom, Hessel, Matteo, van Hasselt, Hado, Lanctot, Marc, and
  de~Freitas, Nando.
\newblock Dueling network architectures for deep reinforcement learning.
\newblock In \emph{Proceedings of The 33rd International Conference on Machine
  Learning}, pp.\  1995--2003, 2016.

\bibitem[Wilcoxon(1945)]{wilcoxon1945individual}
Wilcoxon, Frank.
\newblock Individual comparisons by ranking methods.
\newblock \emph{Biometrics bulletin}, 1\penalty0 (6):\penalty0 80--83, 1945.

\bibitem[Zahavy et~al.(2016)Zahavy, Ben-Zrihem, and Mannor]{zahavy2016graying}
Zahavy, Tom, Ben-Zrihem, Nir, and Mannor, Shie.
\newblock Graying the black box: Understanding dqns.
\newblock In \emph{Proceedings of The 33rd International Conference on Machine
  Learning}, pp.\  1899--1908, 2016.

\end{thebibliography}

\newpage
\appendix
\section{Adding Regularization to LSTD-Q}
\label{sec:lstdq_reg}
For LSTD-Q, regularization cannot be applied directly since the algorithm is finding a fixed-point and not solving a LS problem. To overcome this obstacle, we augment the fixed point function of the LSTD-Q algorithm to include a regularization term based on \citep{kolter2009regularization}: 
\begin{equation}
\label{eq:aug}
f(w) = \argmin _u \| \phi u - \Pi T^* \phi w \| + \lambda g(u) \enspace ,
\end{equation}
where $\Pi$ stands for the linear projection, $T^*$ for the Bellman optimality operator and $g(u)$ is the regularization function. Once the augmented problem is solved, the solution to the regularized LSTD-Q problem is given by $w=f(w)$. This derivation results in the same solution for LSTD-Q as was obtained for FQI (Equation \ref{eq:fqi}). In the special case where $\mu =0,$ we get the $L_2$ regularized solution of \cite{kolter2009regularization}. 

\section{LS-DQN Algorithm}

\begin{figure}[h]
\centering
\includegraphics[width=\textwidth]{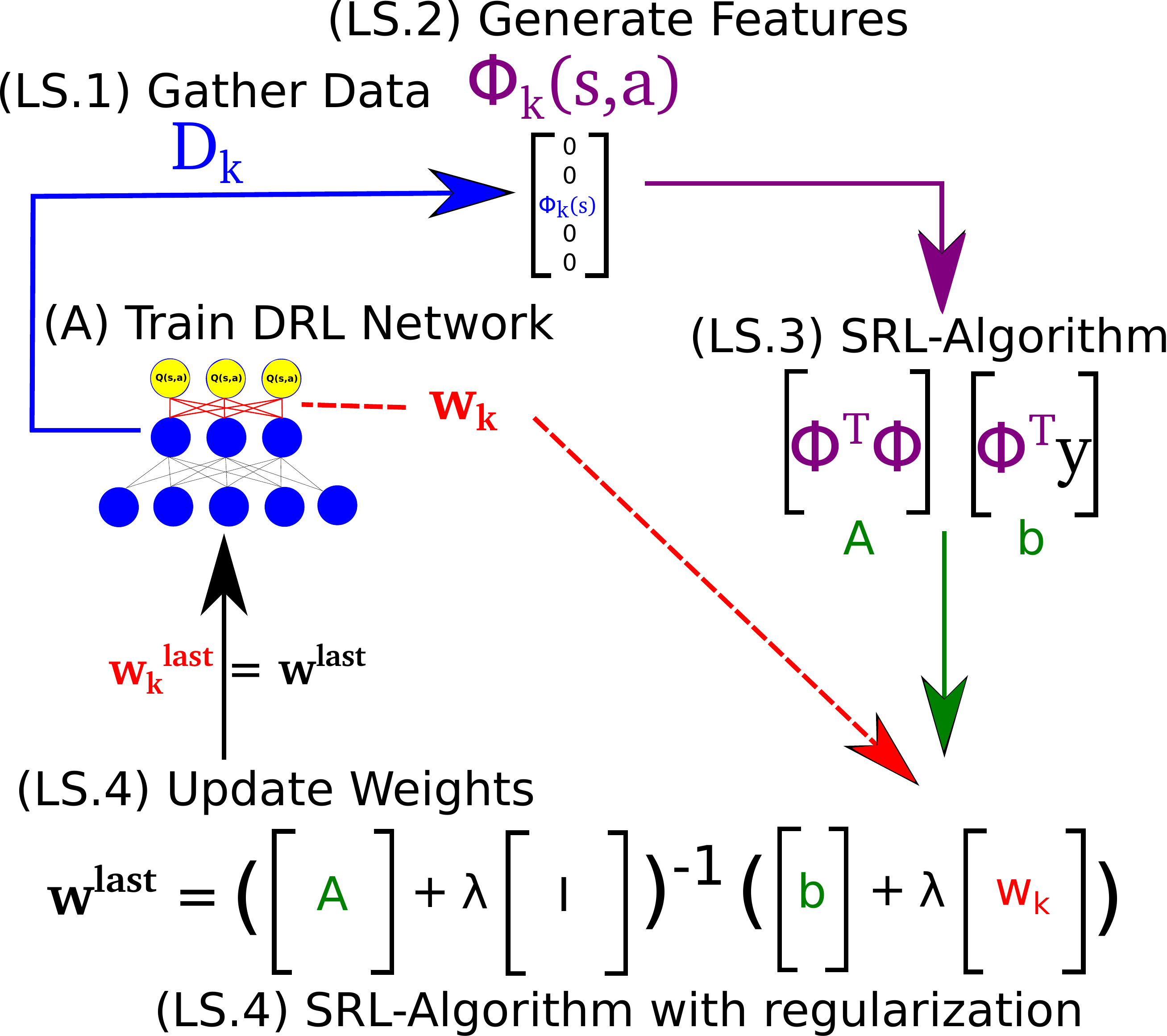}
\caption{An overview of the LS-DQN algorithm.}
\label{alg:lsdqn}
\end{figure}

Figure \ref{alg:lsdqn} provides an overview of the LS-DQN algorithm described in the main paper. The DNN agent is trained for $N_{DRL}$ steps (A). The weights of the last hidden layer are denoted $w_k$. Data is then gathered (LS.1) from the agent's experience replay and features are generated (LS.2). An SRL-Algorithm is applied to the generated features (LS.3) which includes a regularized Bayesian prior weight update (LS.4). Note that the weights $w_k$ are used as the prior. The weights of the last hidden layer are then replaced by the SRL output $w^{last}$ and this process is repeated.

\newpage
\section{Results for SRL Algorithms with High Dimensional Observations}
We present the average scores (averaged over $20$ roll-outs) at different epochs, for both the original DQN and after relearning the last layer using LSTD-Q, for different regularization coefficients.

\textbf{Breakout}

\begin{table}[H]
\caption{Average scores on the different epochs as a function of regularization coefficients}
\begin{center}
\begin{small}
\begin{tabular}{| l || >{\centering\arraybackslash}m{0.045\textwidth} | >{\centering\arraybackslash}m{0.045\textwidth} | >{\centering\arraybackslash}m{0.045\textwidth} | >{\centering\arraybackslash}m{0.045\textwidth} | >{\centering\arraybackslash}m{0.045\textwidth} | >{\centering\arraybackslash}m{0.045\textwidth} | >{\centering\arraybackslash}m{0.045\textwidth} | >{\centering\arraybackslash}m{0.045\textwidth} | >{\centering\arraybackslash}m{0.045\textwidth} | >{\centering\arraybackslash}m{0.045\textwidth} | >{\centering\arraybackslash}m{0.045\textwidth} |} 
\hline
\backslashbox{Epoch}{$\lambda$} & $10^{2}$ & $10^{1}$ & $10^{0}$ & $10^{-1}$ & $10^{-2}$ & $10^{-3}$ & $10^{-4}$ & $10^{-5}$ & $10^{-6}$ & $10^{-7}$ & DQN  \\ \hline \hline
Epoch $1$ & 54 & 49	 & 48 & 44 & 53 & 49 & 48 & 50 & 28 & 30 & 46\\
Epoch $2$ & 207 & 189 & 196 & 193 & 64 & 30 & 18 & 4 & 9 & 5 & 171\\
Epoch $3$ & 238 & 247 & 314 & 284 & 277 & 254 & 270 & 232 & 225 &194 & 271\\
Epoch $4$ & 238 & 271 & 289 & 249 & 207 & 201 & 291 & 326 & 274 &304 & 212\\
Epoch $5$ & 265 & 311 & 322 & 315 & 208 & 109 & 175 & 36 & 14 &48 & 292\\
Epoch $6$ & 299 & 331 & 327 & 328 & 259 & 150 & 248 & 227 & 281 &245 & 164\\
Epoch $7$ & 332 & 335 & 350 & 266 & 128 & 67 & 145 & 249 & 291 &214 & 325\\
Epoch $8$ & 361 & 352 & 343 & 262 & 204 & 65 & 270 & 309 & 287 &304 & 324\\
Epoch $9$ & 294 & 291 & 323 & 319 & 101 & 85 & 224 & 276 & 347 &340 & 350\\
Epoch $10$ & 186 & 297 & 256 & 263 & 243 & 236 & 349 & 323 & 333 &333 & 165\\
Epoch $11$ & 241 & 277 & 290 & 140 & 79 & 111 & 338 & 335 & 330 &315 & 233\\
Epoch $12$ & 328 & 336 & 327 & 352 & 226 & 208 & 337 & 374 & 354 &377 & 302\\
Epoch $13$ & 343 & 305 & 247 & 308 & 62 & 112 & 338 & 342 & 305 &344 & 316\\
Epoch $14$ & 278 & 294 & 259 & 273 & 156 & 198 & 320 & 355 & 350 & 346& 306\\
Epoch $15$ & 312 & 327 & 282 & 292 & 161 & 141 & 321 & 381 & 368 &367 & 252\\
Epoch $16$ & 186 & 160 & 283 & 273 & 170 & 225 & 370 & 314 & 325 &324 & 114\\ \hline
\end{tabular}
\label{table:regularization_breakout}
\end{small}
\end{center}
\vspace{-0.5cm}
\end{table}

\textbf{Qbert}

\begin{table}[H]
\caption{Average scores on the different epochs as a function of regularization coefficients}
\begin{center}
\begin{small}
\begin{tabular}{| l || >{\centering\arraybackslash}m{0.045\textwidth} | >{\centering\arraybackslash}m{0.045\textwidth} | >{\centering\arraybackslash}m{0.045\textwidth} | >{\centering\arraybackslash}m{0.045\textwidth} | >{\centering\arraybackslash}m{0.045\textwidth} | >{\centering\arraybackslash}m{0.045\textwidth} | >{\centering\arraybackslash}m{0.045\textwidth} | >{\centering\arraybackslash}m{0.045\textwidth} | >{\centering\arraybackslash}m{0.045\textwidth} | >{\centering\arraybackslash}m{0.045\textwidth} | >{\centering\arraybackslash}m{0.045\textwidth} |} 
\hline
\backslashbox{Epoch}{$\lambda$} & $10^{2}$ & $10^{1}$ & $10^{0}$ & $10^{-1}$ & $10^{-2}$ & $10^{-3}$ & $10^{-4}$ & $10^{-5}$ & $10^{-6}$ & $10^{-7}$ & DQN  \\ \hline \hline
Epoch $1$ &3470  &3070  &2163  &1998  &1599  &2078  &964  &629  &831  &484  &2978\\
Epoch $2$ &2794  &1853  &2196  &2565  &3839  &3558  &1376  &2123  &1728  &2388  &2060\\
Epoch $3$ &4253  &4188  &4579  &4034  &4031  &2239  &561  &691  &824  &570  &4148\\
Epoch $4$ &2789  &2489  &2536  &2750  &3435  &5214  &2730  &2303  &1356  &594  &1878\\
Epoch $5$ &6426  &6831  &7480  &6703  &3419  &3335  &4205  &3519  &4673  &5231  &7410\\
Epoch $6$ &8480  &7265  &7950  &5300  &4978  &4178  &4533  &6005  &6133  &4829  &8356\\
Epoch $7$ &8176  &9036  &8635  &7774  &7269  &7428  &6196  &3030  &3246  &2343  &8643\\
Epoch $8$ &9104  &10340  &9935  &7293  &7689  &7343  &6728  &2913  &3299  &1473  &9315\\
Epoch $9$ &9274  &10288  &9115  &7508  &6660  &7800  &120  &8133  &4880  &5018  &8156\\
Epoch $10$ &10523  &7245  &9704  &7949  &8640  &7794  &2663  &8905  &10044  &7585  &12584\\
Epoch $11$ &10821  &11510  &9971  &7064  &6836  &9908  &1020  &11868  &9940  &11138  &10290\\
Epoch $12$ &7291  &10134  &7583  &6673  &7815  &9028  &5564  &8893  &8649  &6748  &7438\\
Epoch $13$ &12365  &12220  &13103  &11868  &11531  &10091  &2753  &10804  &8216  &8835  &13054\\
Epoch $14$ &11686  &11085  &10338  &10811  &8386  &9580  &2980  &6469  &6435  &6071  &10249\\
Epoch $15$ &11228  &12841  &13696  &10971  &5820  &10148  &7524  &11959  &9270  &6949  &11630\\
Epoch $16$ &11643  &12489  &13468  &11773  &8191  &8976  &198  &7284  &7598  &5649  &12923\\ \hline
\end{tabular}
\label{table:regularization_qbert}
\end{small}
\end{center}
\vspace{-0.5cm}
\end{table}

%



\section{Results for Ablative Analysis}
\label{sec:ablative}
We used the implementation of ADAM from the \verb+optim+ package for torch that can be found at \url{https://github.com/torch/optim/blob/master/adam.lua}. We used the default hyperparameters (except for the learning rate): learningRate$=0.00025$, learningRateDecay$=0$, beta1$=0.9$, beta2$=0.999$, epsilon$=1$e$-8$, and weightDecay$=0$. For solutions that use the prior, we set $\lambda=1$.

Figure~\ref{fig:score_adam_e1} depicts the offset of the average scores from the DQN's scores, after one iteration of the ADAM algorithm:

\begin{figure}[H]
\begin{center}
  \includegraphics[width=\textwidth]{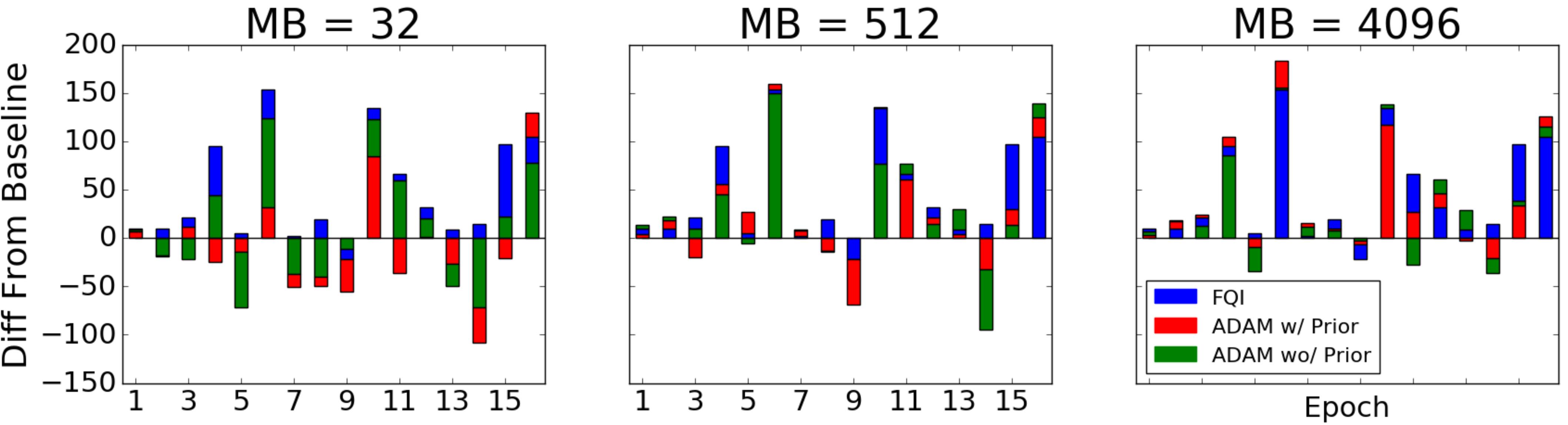}
  \vspace{-0.5cm}
  \caption{Differences of the average scores from DQN compared to ADAM and FQI (with and without priors) for different mini-batches (MB) sizes.}
  \label{fig:score_adam_e1}
\end{center}
\end{figure}

Table~\ref{table:norms} shows the norm of the difference between the different solution weights and the original last layer weights of the DQN (divided by the norm of the DQN's weights for scale), averaged over epochs. Note that MB stands for mini-batch sizes used by the ADAM solver.

\begin{table}[H]
\caption{Norms of the Difference Between solutions Weights}
\begin{center}
\begin{small}
\begin{tabular}{| l || >{\centering\arraybackslash}m{0.1\textwidth} | >{\centering\arraybackslash}m{0.08\textwidth} | >{\centering\arraybackslash}m{0.08\textwidth} | >{\centering\arraybackslash}m{0.08\textwidth} | >{\centering\arraybackslash}m{0.08\textwidth} | >{\centering\arraybackslash}m{0.08\textwidth} | >{\centering\arraybackslash}m{0.08\textwidth} |} 
\hline
 & Batch & MB=32 iter=1 & MB=32 iter=20 & MB=512 iter=1 & MB=512 iter=20 & MB=4096 iter=1 & MB=4096 iter=20  \\ \hline \hline
w/ prior & $\sim$3e-4 & $\sim$3e-3 & $\sim$3e-3 & $\sim$2e-3 & $\sim$2e-3 & $\sim$1.7e-3 & $\sim$1.8e-3\\ \hline
wo/ prior & & $\sim$3.8e-2 & $\sim$2.7e-1 & $\sim$1.3e-2 & $\sim$1.2e-1 & $\sim$5e-3 & $\sim$5e-2\\ \hline
\end{tabular}
\label{table:norms}
\end{small}
\end{center}
\vspace{-0.5cm}
\end{table}

\section{Feature augmentation}
\label{sec:feature_aug}
The LS-DQN algorithm requires a function $\Phi\left(s,a \right)$ that creates features (Algorithm \ref{alg:my_alg}, Line 9) for a dataset $D$ using the current value-based DRL network. Notice that for most value-based DRL networks (e.g. DQN and DDQN), the DRL features (output of the last hidden layer) are a function of the state and not a function of the action. On the other hand, the FQI and LSTDQ algorithms require features that are a function of both state and action. We, therefore, augment the DRL features to be a function of the action in the following manner.  Denote by $\phi\left(s\right) \in \mathbb{R}^f$ the output of the last hidden layer in the DRL network (where $f$ is the number of neurons in this layer). We define $\Phi\left(s,a\right) \in \mathbb{R}^{f|A|}$  to be $\phi\left(s\right)$ on a subset of indices that belongs to action $a$ and zero otherwise, where $|A|$ refers to the size of the action space.

Note that in practice, DQN and DDQN maintain an ER, and we create features for all the states in the ER. A more computationally efficient approach would be to store the features in the ER after the DRL agent visits them, makes a forward propagation (and compute features) and store them in the ER. However, SRL algorithms work only with features that are fixed over time. Therefore, we generate new features with the current DRL network. 

\end{document}